\documentclass[10pt,twocolumn,letterpaper]{article}

\usepackage{icb}
\usepackage{times}
\usepackage{epsfig}
\usepackage{graphicx}
\usepackage{amsmath}
\usepackage{amssymb}

\usepackage{multirow}

\icbfinalcopy 

\ificbfinal\fi
\begin{document}

\title{Alignment Free and Distortion Robust Iris Recognition}

\author{Min Ren\textsuperscript{1 2 3}, Caiyong Wang\textsuperscript{1 2 3}, Yunlong Wang\textsuperscript{2 3}, Zhenan Sun\textsuperscript{2 3}, Tieniu Tan\textsuperscript{2 3}\\
\textsuperscript{1}University of Chinese Academy of Sciences, \textsuperscript{2}CRIPAC CASIA, \textsuperscript{3}NLPR CASIA, Beijing, P.R. China\\
\{min.ren, caiyong.wang, yunlong.wang\}@cripac.ia.ac.cn, \{znsun, tnt\}@nlpr.ia.ac.cn
}

\maketitle
\thispagestyle{empty}

\begin{abstract}
Iris recognition is a reliable personal identification method but there is still much room to improve its accuracy especially in less-constrained situations. For example, free movement of head pose may cause large rotation difference between iris images. And illumination variations may cause irregular distortion of iris texture. To match intra-class iris images with head rotation robustly, the existing solutions usually need a precise alignment operation by exhaustive search within a determined range in iris image preprosessing or brute force searching the minimum Hamming distance in iris feature matching. In the wild, iris rotation is of much greater uncertainty than that in constrained situations and exhaustive search within a determined range is impracticable. This paper presents a unified feature-level solution to both alignment free and distortion robust iris recognition in the wild. A new deep learning based method named Alignment Free Iris Network (AFINet) is proposed, which uses a trainable VLAD (Vector of Locally Aggregated Descriptors) encoder called NetVLAD~\cite{Arandjelovic2017NetVLAD} to decouple the correlations between local representations and their spatial positions. And deformable convolution~\cite{Dai2017Deformable} is used to overcome iris texture distortion by dense adaptive sampling. The results of extensive experiments on three public iris image databases and the simulated degradation databases show that AFINet significantly outperforms state-of-art iris recognition methods.

\end{abstract}

\section{Introduction}
Iris recognition technology is an attractive topic of biometrics and has been widely applied in person identification because of the rich and unique texture information human iris can provide. Since the first successful algorithm for iris recognition~\cite{Daugman1993High}, great improvements have been made so far to solve high-quality iris image matching under constrained conditions.

\begin{figure}
\begin{center}
\includegraphics[width=0.4\textwidth]{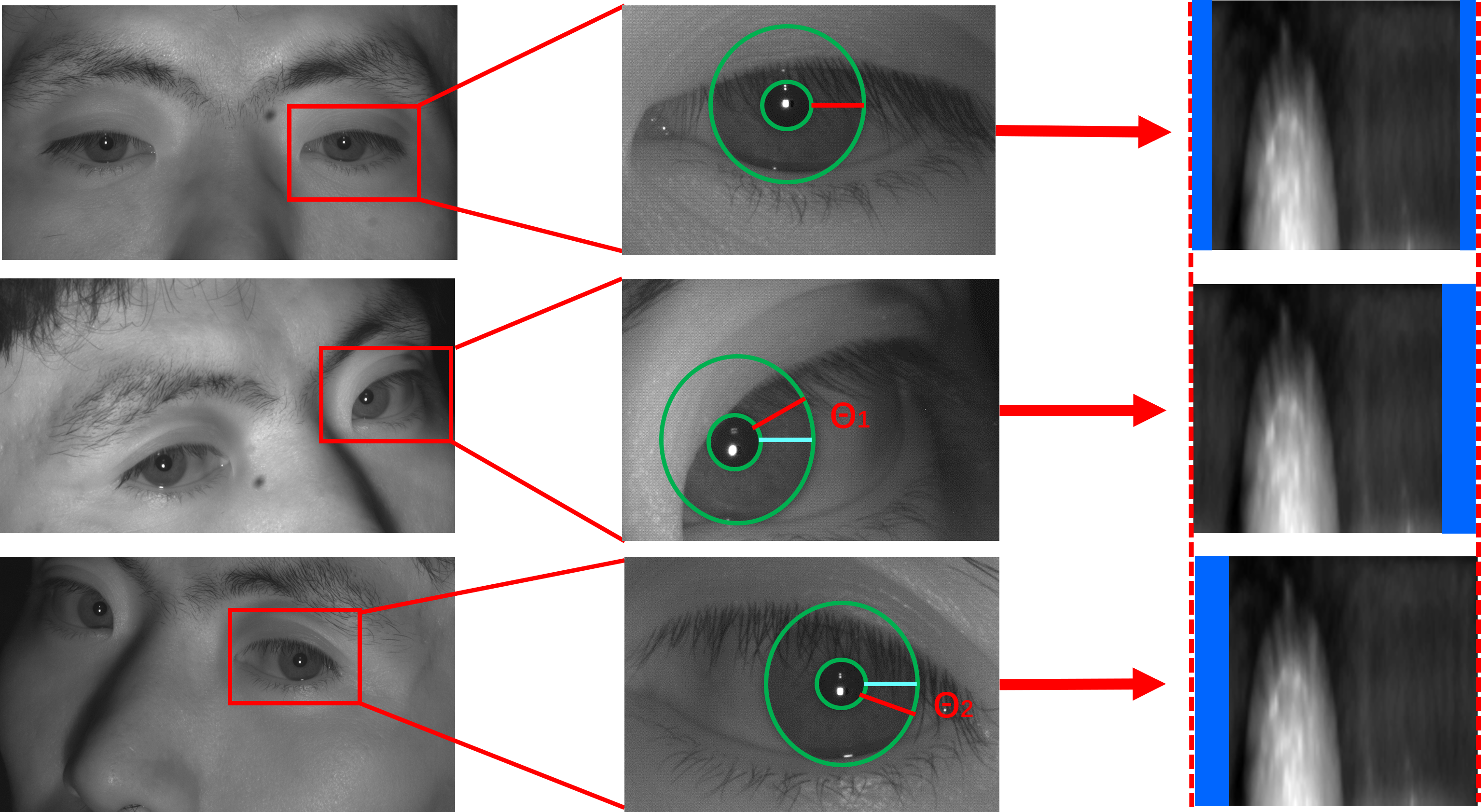}
\end{center}
   \caption{The first row shows the normal iris sample under constrained situations. The rest two rows show the situation without constrain: a glimpse while walking at a distance. The rotation shift of iris may be much greater than constrained situations. The rotation shift of iris becomes horizontally shift in the normalized iris images as shown in the right column.}
\label{fig:show}
\end{figure}

However, iris recognition in the wild environments, especially where subjects are standing at a distance and not cooperative, is still challenging. Standing distance, gaze direction and shooting angle exhibit significant variations as shown in Figure~\ref{fig:show}. Two severe obstacles for accurate iris recognition arise: (i) The rotation of iris is of much greater uncertainty than that in constrained situations, which can occasionally reach $45^\circ$. (ii) Distortion of iris texture is more serious than constrained situations because of the complicated illumination in the wild. 

Currently, almost all existing methods for iris recognition, including handcraft-designed methods and deep learning based methods, need to align iris samples by a cumbersome process during both training and testing~\cite{Daugman1993High}~\cite{Zhenan2009Ordinal}~\cite{Zhang2018Deep}~\cite{Zhao2017Towards}: shift the normalized iris image horizontally within a small and determined range to exhaustively search the best position for matching. However, this process only works well in constrained situations. While in the wild environments, where the range of iris rotation is not determined, the method of exhaustion is impracticable to align iris samples beyond the searching range. Iris texture may have irregular distortions under different illumination conditions. And these distortions are difficult to be precisely modeled and corrected in preprocessing step. Traditional solution is to use robust features against iris pattern distortions. However considerable high rates of false rejection are still reported in recent iris evaluations because the difference between intra-class iris patterns can not be perfectly resolved using existing robust iris features. So misalignment and distortion are important factors affecting iris recognition accuracy in the wild and they are open problems in iris biometrcis needing more investigation.

A novel deep learning based framework named Alignment Free Iris Network (AFINet) is proposed in this paper to solve both alignment free and distortion robust iris recognition. Our model consists of three components: local feature extractor, spatial-invariant encoder and classifier. Deformable convolution~\cite{Dai2017Deformable} is incorporated into convolutional neural network for dense adaptive sampling to overcome the distortion of iris texture. Inspired by NetVLAD~\cite{Arandjelovic2017NetVLAD}, which is a trainable VLAD encoder, we decouple the correlations between local representations and their spatial positions. Full connected layers are adopted as classifier.

The main contributions of this paper are summarized as follows: (i) a new deep learning based framework is proposed to solve alignment free and distortion robust iris recognition, (ii) results of extensive experiments show that our method remarkably outperforms several state-of-art methods in various settings.

\section{Related Works}
Numerous of iris recognition methods were proposed since the successful introduction of Gabor filters in 1993 ~\cite{Daugman1993High}. One of the challenge problems for iris recognition is iris feature extraction. 1D log-Gabor filters was proposed~\cite{L2003} for efficient iris feature extraction. Discrete cosine transforms (DCT) was exploited~\cite{Monro2007} to produce binary iris features by analyzing frequency information of the iris image. 2D discrete Fourier transforms (DFT) proposed in~\cite{Kazuyuki2008} also attempted to match iris images by frequency analysis. An universal framework of iris feature extraction named Ordinal Measures (OMs) was proposed by Sun \emph{et al.}~\cite{Zhenan2009Ordinal}, which extract spatial texture features for matching. Adaptive bloom filters were applied for alignment-free cancelable iris biometric~\cite{C2013Alignment}. However, this method was a framework tailored for feature transformation rather than feature extraction, so it was limited in general iris feature extraction methods.

Deep learning based methods for iris recognition have been developed in recent years. Liu \emph{et al.} proposed DeepIris in~\cite{Liu2016DeepIris} for heterogeneous iris matching by pairwise filter learning. FCN based framework was investigated in~\cite{Zhao2017Towards} for iris recognition and a method named UniNet was proposed. A CNN based method named MaxoutCNNs was proposed in~\cite{Zhang2018Deep} for iris and periocular recognition on mobile devices.

Overall these methods need cumbersome alignment operation before iris feature extraction and can not handle the distortion of iris texture well.

\begin{figure*}[htb]
\begin{center}
\includegraphics[width=.9\textwidth]{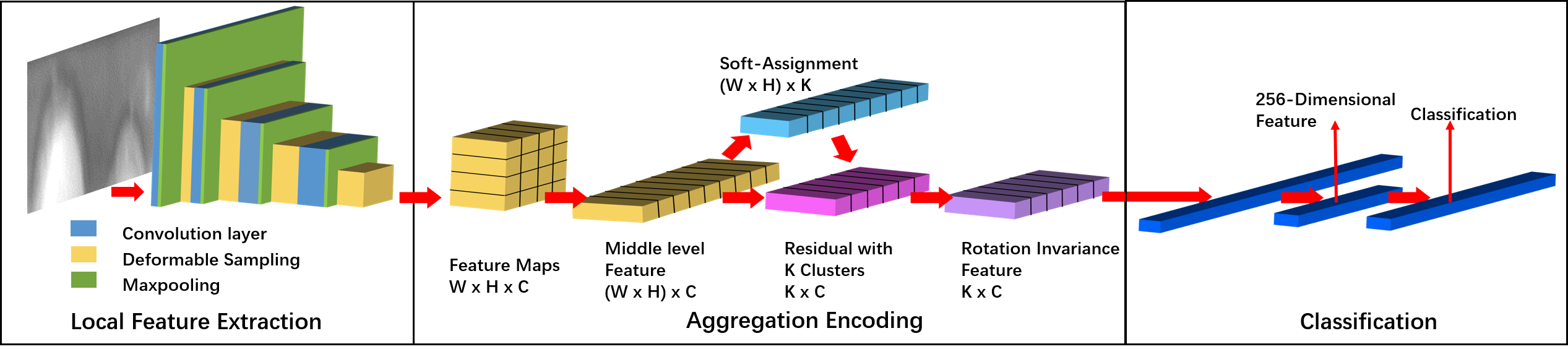}
\end{center}
   \caption{Framework of Alignment Free Iris Network consist of three stages: local feature extractor which overcomes iris texture distortion by dense adaptive sampling while extracts local features from iris images, aggregation encoder which aggregates local features to rotation insensitive representations and classifier which reduces the feature dimension before classification.}
\label{fig:framework}
\end{figure*}

\section{Alignment Free Iris Networks}
This section describes the proposed deep learning framework for alignment free and distortion robust iris recognition. Iris rotation on 2D image planes, which presents as horizontal shifting in the normalized iris image, is much more serious in the wild environments. To solve this problem, the proposed framework decouples the correlations between local representations and their spatial positions to generate rotation-insensitive representation. Iris texture distortions, usually caused by varying illuminations, are resolved by dense adaptive sampling. AFINet can be trained in an end-to-end fashion for alignment free and distortion robust iris recognition.

\subsection{Iris Image Preprocessing}
For all of the iris images this paper, we applied the same preprocessing procedure which contains four steps: 1) Eye detection using Haar-like Adaboost detectors~\cite{Viola2004Robust}. 2) Pupillary and iris boundaries localization by method of ~\cite{Zhaofeng2009Toward}. 3) Annular iris image normalized to a rectangle image with $128\times 128$ resolution by rubber sheet model~\cite{Daugman1993High}. 4) Pixel intensity normalization by the mean and standard  deviation of each database.

\begin{table}
\begin{center}
\caption{Layer configurations of feature extractor.}
\label{t:e}
\begin{tabular}{p{55pt}|p{30pt}|p{25pt}|p{40pt}}
\hline
{\bf Layer}  & {\bf Kernel size} & \bf{Stride}  & \bf{Output channels}\\
\hline\hline
Conv1  & $9\times 9$ & 1  & 48\\
\hline
Maxpooling1 & $2\times 2$ & 2  & -\\
\hline
DeformConv1  & $3\times 3$ & 1  & -\\
\hline
Conv2 & $5\times 5$ & 1  & 96\\
\hline
 Maxpooling2 & $2\times 2$ & 2  & -\\
\hline
DeformConv2  & $3\times 3$ & 1  & -\\
\hline
Conv3  & $5\times 5$ & 1  & 128\\
\hline
 Maxpooling3 & $2\times 2$ & 2  & -\\
\hline
 DeformConv3  & $3\times 3$ & 1  & -\\
\hline
Conv4 & $4\times 4$ & 1 &  192\\
\hline
 Maxpooling4 & $2\times 2$ & 2  & -\\
\hline
DeformConv4  & $3\times 3$ & 1  & -\\
\hline
\end{tabular}
\end{center}
\end{table}

\subsection{Network Architecture}
The proposed network contains three components as shown in Figure \ref{fig:framework}: local feature extractor, aggregation encoder and classifier. 

 Superposition of convolution layers is able to extract discriminative representation from raw data~\cite{Krizhevsky2012ImageNet}\cite{L1998Gradient}. This kind of architecture is adopted as the first component which contains four convolution layers. The configurations of the local feature extractor are shown in Table~\ref{t:e}. The structure of convolutional layers is based on MaxoutCNNs~\cite{Zhang2018Deep}. Deformable convolution~\cite{Dai2017Deformable} is adopted to overcome iris distortion which will be discussed in next section. All of the convolution layers adopt Maxout activation function and all of the deformable convolution layers use linear activation function. 
 
The local features need to be aggregated to rotation-insensitive  representations. Actually, feature aggregation is a significant research topic, and some classical methods have been proposed, including bag-of-visual-words~\cite{Philbin2007Object}, VLAD~\cite{Jegou2010Aggregating}\cite{Arandjelovic2013All} and Fisher vector~\cite{Herv2012Aggregating}\cite{Perronnin2010Large}. We make use of a trainable aggregation strategy, NetVLAD~\cite{Arandjelovic2017NetVLAD}, a variant of the VLAD to decouple the correlations between local representations and their spatial positions, and more details will be shown later.

Two fully connected layers play the parts of dimension reduction and classification. The first fully connected layer reduces the feature dimension to 256 and its output serves as feature for matching. The output of second fully connected layers is used for loss computation.

\subsection{Deformable Sampling}
Iris texture may have irregular distortions under complicated illumination. Although iris images are normalized to a rectangle image with a fixed resolution, the distortions are difficult to be corrected and this problem is still a important factor affecting iris recognition accuracy in the wild.

Deformable convolution provides an effective approach to tackle this problem by dense adaptive sampling. The outputs of deformable sampling layers are adaptive sampled from the inputs. The adaptive sampling operation is trained according to supervised signals. This operation can be viewed as a nonlinear function which adaptively corrects the iris texture distortion and maps the intra-class iris samples to close positions of feature space.

\begin{figure}[htb]
\begin{center}
\includegraphics[width=.4\textwidth]{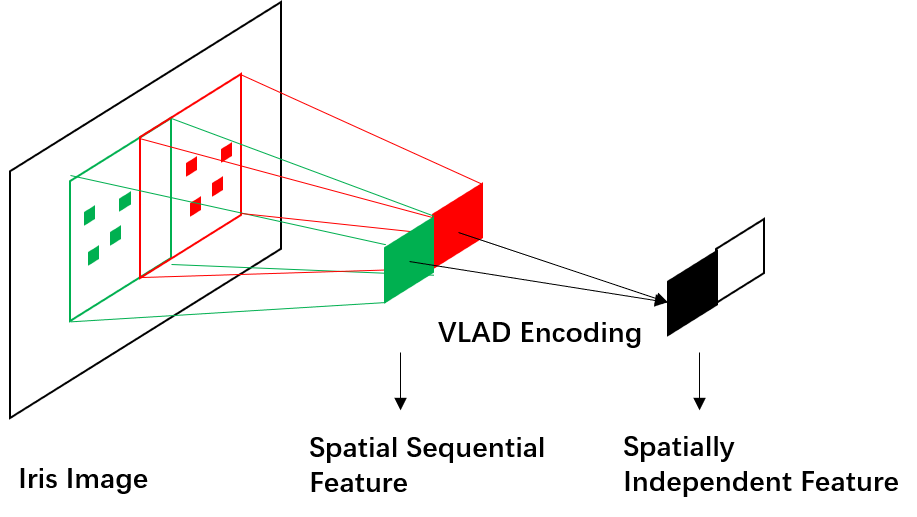}
\end{center}
   \caption{Spatially independent encoding. The similar local patterns of iris image at different positions will activate different nodes of the feature map. After spatially independent encoding, the similar local patterns activate the same nodes and the representation will be rotation insensitive.}
\label{fig:vlad}
\end{figure}

\begin{figure*}[htb]
\centering
\includegraphics[width=.95\textwidth]{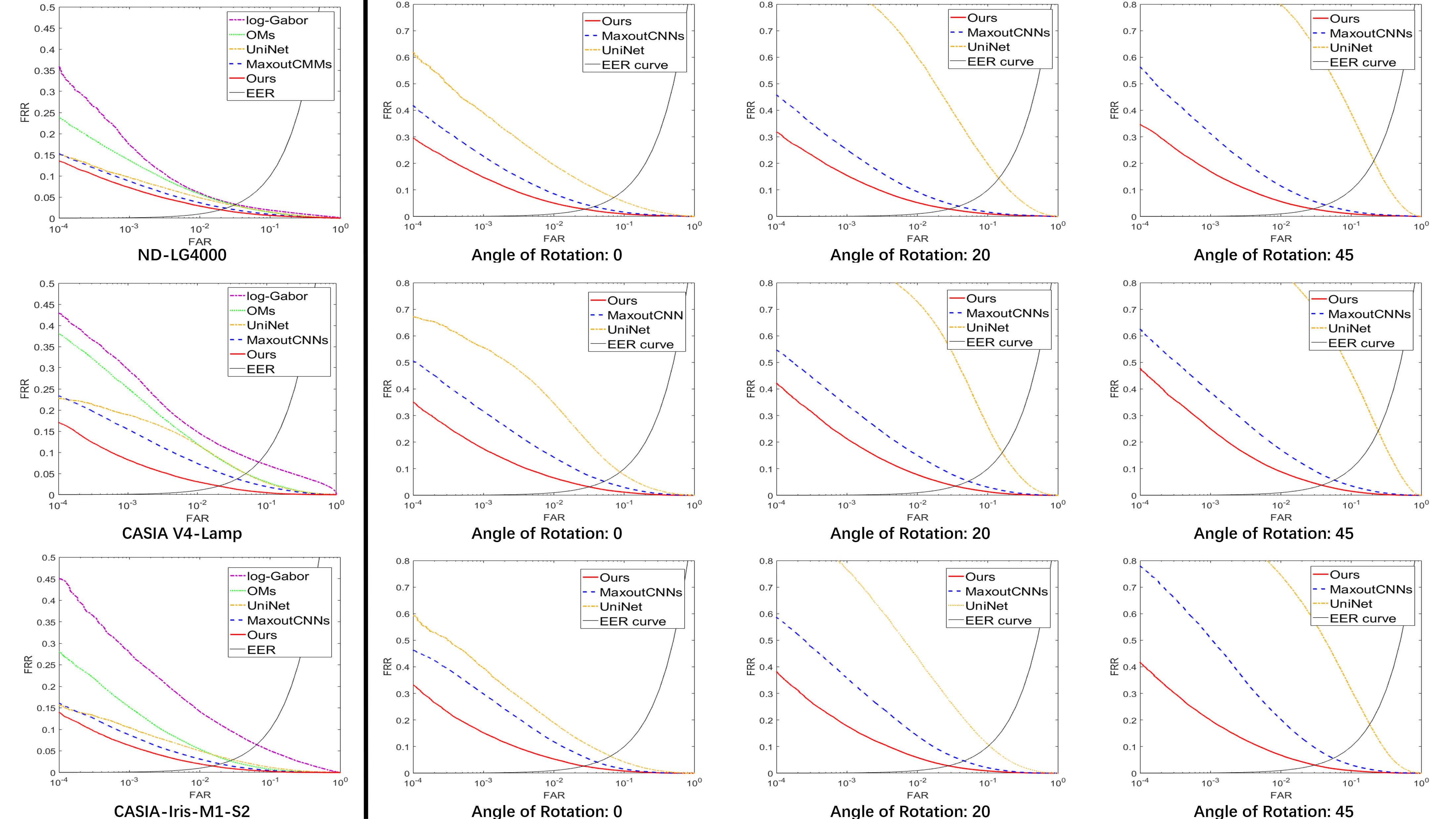}
\caption{ROC curves of the experiments. The first row is ROC curves on ND-CrossSensor-Iris-2013 Dataset - LG4000, the second row is ROC curves on CASIA Iris Image Database V4 - Lamp, the third row is ROC curves on CASIA-Iris-M1-S2. The first column from left shows the results of experiments without iris rotation. The rest three columns shows the results of experiments with iris rotation. The second column from left shows the results under $0^\circ$ iris rotation, the third column from left shows the results under $20^\circ$ iris rotation and the right-most column shows the results under $45^\circ$ iris rotation. Note that the settings of experiments without iris rotation is different from the experiments under $0^\circ$ iris rotation. The models in experiments without iris rotation were trained on original databases, and the models in experiments under $0^\circ$ iris rotation were trained on the degraded databases.}
\label{fig:roc}
\end{figure*}

\subsection{Aggregation Encoding}
The key component of feature aggregation is the trainable VLAD encoder. Feature maps of layer DeformConv4 with dimension $W\times H\times C$ are re-organized as $W\times H$ C-dimension vectors. Each of these vector represents a special patch of the iris image, and they are spatially correlated. The similar local patterns of iris image at different positions are expected to be decoupled with their positions and aggregated into the same representation which means activate the same nodes of network as shown in Figure~\ref{fig:vlad}. So, these vectors are expected to be encoded in a spatially independent way for rotation insensitive representation.
NetVLAD is a differentiable spatially independent method, and it has been shown to outperform average and max pooling for vector aggregation~\cite{Arandjelovic2017NetVLAD}, which makes it perfectly suited for our task.

We provide a brief overview of NetVLAD in this paper (full details are available in ~\cite{Arandjelovic2017NetVLAD}). Formally, $W\times H$ C-dimensional vectors $\{ v_i\}$ are represented by their residuals with $K$ cluster centers. Each $v_i$ are assignmented to cluster centers in a soft way. Spatially independent feature, which means rotation insensitive feature in our task, are produced according to equation~\ref{equation:vlad} (the dimension is $K\times C$):
\begin{equation}
 V(k,~j) = \sum_{i=1}^{W\times H} \frac{e^{W_k^Tv_i+b_k}}{\sum_{k'}e^{W_{k'}^Tv_i+b_{k'}}}(v_i(j) - c_k(j))
\label{equation:vlad}
\end{equation}
where $\{W_k\}$, $\{b_k\}$, $\{c_k\}$ are trainable parameters, with $k\in \{1,~2,~...,~K\},~j\in \{1,~2,~...,~C\}$. $\{c_k\}$ is the cluster centers and the second term of equation~\ref{equation:vlad} represents residuals between middle level feature $\{v_i\}$ and $\{c_k\}$. The first term of equation~\ref{equation:vlad} represents the soft-assignment weight of $\{v_i\}$ and it can be implemented by $1\times 1$ convolution followed by softmax. The final rotation in variant feature is obtained by performing L2 normalization.

\subsection{Implementation Details}
Network initialization and training details are provided in this section.

The cluster centers of NetVLAD are initialized in the same way of~\cite{Arandjelovic2017NetVLAD}: clustering the local vectors with k-means into $K$ clusters. The number of clusters is 25. The other parameters of the network are initialized by sampling from a standardized normal distribution.

The first component of the network used as a local feature extractor is pre-trained on each iris database. After that, the VLAD encoder is adopted for training. Cross entropy loss is used for measuring loss of classification. The initial learning rate of local feature extractor is $10^{-4}$. And the initial learning rate of VLAD encoder and fully connected layers is $10^{-2}$. The learning rates of whole network are divided by 10 until validation error stagnates. The weight decay and momentum, which are $10^{-4}$ and 0.9 respectively, are fixed during training.

\section{Experiments and Results}
Extensive experiments are conducted and results are shown in this section to evaluate the performance of proposed method.

\begin{table*}
\begin{center}
\caption{False reject rates (FRR) at $10^{-4}$ false accept rates (FAR) and equal error rates (EER) of experiments without rotation.}
\label{t:rn}
\setlength{\tabcolsep}{6mm}{
\begin{tabular}{c|c|c|c|c|c|c}
\hline
\multirow{2}*{~} & \multicolumn{2}{|c|}{\bf{ND-LG4000}} & \multicolumn{2}{|c|}{\bf{CASIA V4-Lamp}} & \multicolumn{2}{|c}{\bf{CASIA-Iris-M1-S2}} \\
 \cline{2-7}
& FRR & EER & FRR & EER & FRR & EER \\
\hline
log-Gabor & 35.90\% & 3.25\% & 42.93\% & 7.69\% & 45.03\% & 6.57\% \\
\hline
OMs & 23.88\% & 3.12\% & 38.07\% & 4.99\% & 28.02\% & 2.69\% \\
\hline
UniNet & 15.23\% & 3.01\% & 22.76\% & 4.96\% & 15.60\% & 2.95\% \\
\hline
MaxouCNNs & 15.27\% & 2.36\% & 23.36\% & 3.73\% & 16.08\% & 1.97\% \\
\hline
Ours & \bf{13.55\%} & \bf{1.98\%} & \bf{17.13\%} & \bf{2.04\%} & \bf{13.97\%} & \bf{1.51\%} \\
\hline
\end{tabular}}
\end{center}
\end{table*}

\begin{table*}
\begin{center}
\caption{False reject rates (FRR) at $10^{-4}$ false accept rates (FAR) and equal error rates (EER) of experiments with different rotation.}
\label{t:rr}
\setlength{\tabcolsep}{6mm}{
\begin{tabular}{c|c|c|c|c|c|c}
\hline
\multirow{2}*{~} & \multicolumn{2}{|c|}{\bf{ND-LG4000}} & \multicolumn{2}{|c|}{\bf{CASIA V4-Lamp}} & \multicolumn{2}{|c}{\bf{CASIA-Iris-M1-S2}} \\
 \cline{2-7}
& FRR & EER & FRR & EER & FRR & EER \\
\hline
~ & \multicolumn{6}{c}{\bf{Angle difference between sample pairs: $0^\circ$}}\\ 
\hline
UniNet & 61.71\% & 7.13\% & 67.13\% & 8.74\% &59.26\% & 6.39\% \\
\hline
MaxouCNNs & 41.81\% & 3.75\% & 50.56\% & 5.19\% & 46.31\% & 4.12\% \\
\hline
Ours & \bf{29.65\%} & \bf{2.76\%} & \bf{35.13\%} & \bf{3.18\%} & \bf{33.19\%} & \bf{2.67\%} \\
\hline
~ & \multicolumn{6}{c}{\bf{Angle difference between sample pairs: $20^\circ$}}\\ 
\hline
UniNet & 96.03\% & 14.34\% & 97.84\% & 16.02\% &93.20\% & 10.42\% \\
\hline
MaxouCNNs & 45.88\% & 3.87\% & 54.65\% & 5.30\% & 58.65\% & 4.60\% \\
\hline
Ours & \bf{31.82\%} & \bf{2.81\%} & \bf{42.19\%} & \bf{3.47\%} & \bf{38.22\%} & \bf{2.77\%} \\
\hline
~ & \multicolumn{6}{c}{\bf{Angle difference between sample pairs: $45^\circ$}}\\ 
\hline
UniNet & 98.62\% & 20.96\% & 98.94\% & 24.14\% &98.70\% & 18.30\% \\
\hline
MaxouCNNs & 56.36\% &4.32\% & 62.59\% & 5.75\% & 77.91\% & 5.57\% \\
\hline
Ours & \bf{34.80\%} & \bf{2.85\%} & \bf{47.87\%} & \bf{3.80\%} & \bf{41.64\%} & \bf{3.04\%} \\
\hline
\end{tabular}}
\end{center}
\end{table*}

\subsection{Databases}
\textbf{ND CrossSensor Iris 2013 Dataset-LG4000}. This database~\cite{nd} contains 29,986 iris samples from 1,352 classes and is one of the most popular iris database for iris recognition research. The training set of this database consists of 676 classes which are selected randomly. The testing set is the other 676 classes which are not overlapped with the training set. The training set and test set contain 15,012 and 14,974 samples respectively.

\textbf{CASIA Iris Image Database V4-Lamp}. This database ~\cite{lamp} contains 16,212 iris samples from 819 classes. A lamp was turned on/off close to the subject to introduce elastic deformation of iris texture due to pupil expansion and contraction under different illumination conditions. So this database is suitable  for evaluating the ability of the proposed method to deal with iris distortion. The training set of this database contains 409 classes which are selected randomly. The testing set is the rest 410 classes which are not overlapped with the training set. The training set and test set contain 8,115 and 8,097 samples respectively.

\textbf{CASIA-Iris-M1-S2}. This database~\cite{m1s2} includes 6000 images captured from 200 subjects. Iris samples were captured by mobile devices at different distances. Both eyes of subjects are captured, so that 400 classes are contained in this database. The training set of this database contains 200 classes which are selected randomly. The testing set is the other 200 classes which are not overlapped with the training set. Hence, the numbers of samples of training set and testing set are both 6,000.

\subsection{Experiments without Iris Rotation}
Experiments on the three databases were launched firstly. The proposed method was compared with the state-of-art approaches, including log-Gabor filters~\cite{L2003}, Ordinal Measures (OMs)~\cite{Zhenan2009Ordinal}, MaxouCNNs~\cite{Zhang2018Deep} and UniNet~\cite{Zhao2017Towards}. Both handcraft-designed methods and deep learning based methods are contained for comparison.

The results are shown in the first column from left of Figure~\ref{fig:roc} and Table~\ref{t:rn}. Significant improvements from proposed method can be observed on all three databases especially on the CASIA V4-Lamp. Such results suggest that the proposed method is a better feature extractor for iris data. Two-folds reasons are considered. Firstly, our proposed method handles small range of iris rotation which exists in these databases better than others. Secondly, the distortion of iris texture is well overcome because of the deformable sampling and this can be observed more obviously on CASIA V4-Lamp which introduced elastic deformation of iris texture under different illumination conditions.

\subsection{Experiments with Iris Rotation}
To evaluate the performance of proposed method, iris samples of training set were degraded by rotating a random angle, and the angle was sampled from a uniform distribution from $0^\circ$ to $45^\circ$. Three configurations were taken into consideration during testing: $0^\circ$, $20^\circ$ and $45^\circ$ difference between sample pairs for matching (horizontal translations of normalized iris images were applied during implementation according to angles). Handcraft-designed methods were not contained because they can not work without the strategy of exhaustive search.

The results are shown in right three columns of Figure~\ref{fig:roc} and Table~\ref{t:rr}. The proposed method surpasses other methods by a large margin, and the gain in accuracy grow with the increase of angle of rotation. These results suggest that our network does capture rotation insensitive features from the iris images, and the representation of iris image produced by our network is suitable for alignment free and distortion robust iris recognition. Experimental evidence are shown in next section.

\subsection{Experimental Analysis}
\begin{figure}[htb]
\begin{center}
\includegraphics[width=.45\textwidth]{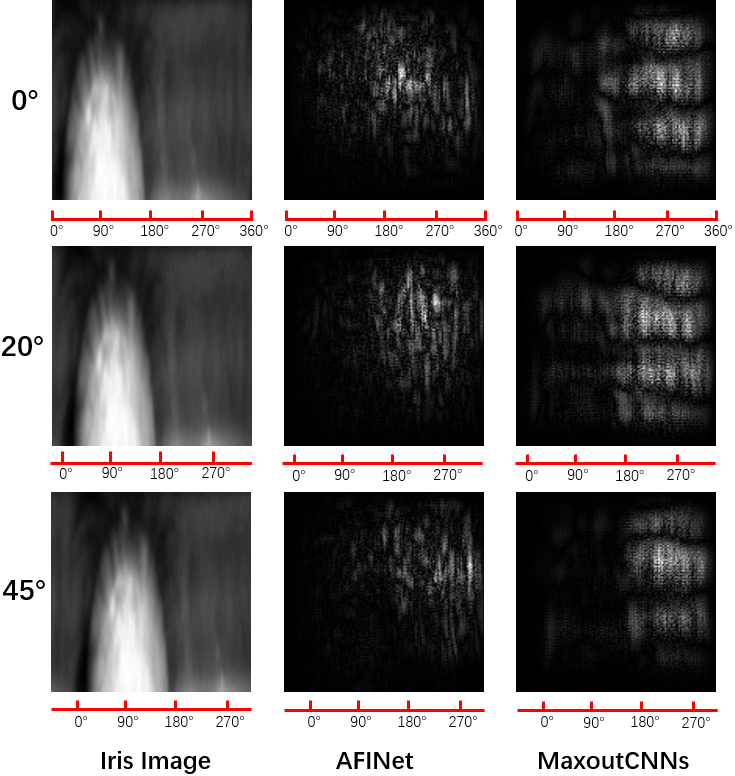}
\end{center}
   \caption{Saliency maps of the same iris sample under different rotation angle. The left column is normalized iris image under three different rotation angles: $0^\circ$, $20^\circ$, $45^\circ$. The middle column is the saliency maps of proposed network (the brighter positions represent higher responses which means more important for classification). The saliency maps of MaxoutCNNs are shown in right column as comparison. }
\label{fig:salmap}
\end{figure}

To demonstrate the capability of capturing the similar local patterns under different rotation angles, saliency maps~\cite{Simonyan2013Deep} were adopted to visualize the important areas of iris image by back-propagation. Saliency maps of the same iris sample under different rotation angle are shown in Figure~\ref{fig:salmap}.

The similar high response positions of saliency map of proposed network move right with the rotation of input iris sample as shown in middle column of Figure~\ref{fig:salmap}. This phenomenon dose not appear in the right column, which is the saliency maps of MaxoutCNNs. It is demonstrated that our network capture the similar local patterns of iris image under different rotation angles. This ability is not observed in MaxoutCNNs which has similar architecture with our  network except the VLAD encoder and deformable sampling.

\section{Conclusion}
In this paper, we analyze the difficulties of iris recognition in the wild and propose a novel framework for alignment free and distortion robust iris recognition. Extensive experiments demonstrate the effectiveness of our model compared to existing methods. Alignment free and distortion robust are open problems for iris recognition deserving further investigation and iris recognition in the wild is still a challenging problem Our future work would focus on robust representation for iris sample matching. 


{\small
\bibliographystyle{ieee}
\bibliography{egbib}
}

\end{document}